\documentclass[10pt,twocolumn,letterpaper]{article}

\usepackage{cvpr}
\usepackage{times}
\usepackage{epsfig}
\usepackage{graphicx}
\usepackage{amsmath}
\usepackage{amssymb}
\usepackage{subfig}
\usepackage{enumitem}
\usepackage{tabularx}
\usepackage[flushleft]{threeparttable} 
\usepackage[export]{adjustbox}
\usepackage{booktabs}
\usepackage[dvipsnames, table]{xcolor}

\usepackage{hyperref}

\hypersetup{
  colorlinks, linkcolor=red
}
\makeatletter
\g@addto@macro{\UrlBreaks}{\UrlOrds}
\makeatother

\newcommand{\bx}{\mathbf{x}}

\newcommand{\bdelta}{\mathbf{\delta}}

\newcommand{\IR}{\mathbb{R}}

\newcommand{\EX}{\mathcal{X}}
\newcommand{\EZ}{\mathcal{Z}}
\newcommand{\EP}{\mathcal{P}}
\newcommand{\ES}{\mathcal{S}} 
\newcommand{\EL}{\mathcal{L}}

\newcommand{\EA}{\mathcal{A}}

\DeclareRobustCommand{\ie}{\textit{i}.\textit{e}.\@\xspace}
\DeclareRobustCommand{\etal}{\textit{et al.}\@\xspace}

\definecolor{myGreen}{rgb}{0.17,0.64,0.37}
\definecolor{myRed}{rgb}{0.95,0.1,0.10}

\newlength{\tempdima}
\newcommand{\rowname}[1]
{\rotatebox{90}{\makebox[\tempdima][c]{\textbf{#1}}}}

\makeatletter
\renewcommand\paragraph{\@startsection{paragraph}{4}{\z@}%
                                      {\parskip}
                                      {-1em}%
                                      {\normalfont\normalsize\bfseries}}
\makeatother

\cvprfinalcopy 

\def\cvprPaperID{10182} 

\ifcvprfinal\fi
\begin{document}

\title{Finding Missing Children: Aging Deep Face Features}

\author{Debayan Deb\\
Michigan State University\\
East Lansing, MI, USA\\
{\tt\small debdebay@msu.edu}
\and
Divyansh Aggarwal\\
Michigan State University\\
East Lansing, MI, USA\\
{\tt\small aggarw49@msu.edu}
\and
Anil K. Jain\\
Michigan State University\\
East Lansing, MI, USA\\
{\tt\small jain@cse.msu.edu}
}


\twocolumn[{%
\renewcommand\twocolumn[1][]{#1}%
\maketitle
 \thispagestyle{empty}
\begin{center}
\vspace{-1.2em}
    \centering
    \footnotesize
    \captionsetup{font=footnotesize}
    \setlength{\fboxsep}{2.2pt}
    \setlength{\fboxrule}{0.2pt}
    \begin{minipage}{0.18\linewidth}
    \fcolorbox{white}{white}{\includegraphics[width=0.95\linewidth]{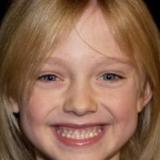}}\\
    \centering {\small7 years old}\vspace{0.5em}
    \fcolorbox{white}{white}{\includegraphics[width=0.95\linewidth]{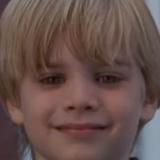}}\\
    \centering {\small 7 years old}
    \end{minipage}\;
    \begin{minipage}{0.18\linewidth}
    \fcolorbox{white}{Green}{\includegraphics[width=0.95\linewidth]{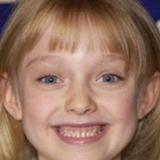}}%
    \\
    \centering {\small8 years old}\vspace{0.5em}
    \fcolorbox{white}{Green}{\includegraphics[width=0.95\linewidth]{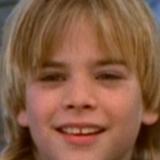}}\\
    \centering {\small11 years old}
    \end{minipage}\;
    \begin{minipage}{0.18\linewidth}
    \fcolorbox{white}{Green}{\includegraphics[width=0.95\linewidth]{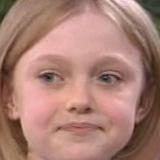}}\\
     \centering {\small10 years old}\vspace{0.5em}
    \fcolorbox{white}{Red}{\includegraphics[width=0.95\linewidth]{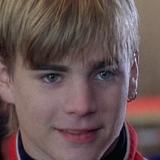}}\\
    \centering {\small17 years old}
    \end{minipage}\;
    \begin{minipage}{0.18\linewidth}
    \fcolorbox{white}{Red}{\includegraphics[width=0.95\linewidth]{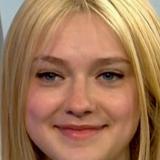}}\\
    \centering {\small16 years old}\vspace{0.5em}
    \fcolorbox{white}{Red}{\includegraphics[width=0.95\linewidth]{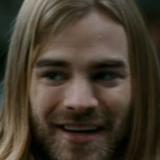}}\\
    \centering {\small25 years old}
    \end{minipage}\;
    \begin{minipage}{0.18\linewidth}
    \fcolorbox{white}{Red}{\includegraphics[width=0.95\linewidth]{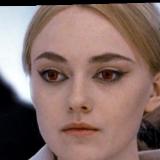}}\\
    \centering {\small18 years old}\vspace{0.5em}
    \fcolorbox{white}{Red}{\includegraphics[width=0.95\linewidth]{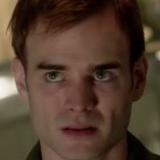}}\\
    \centering {\small28 years old}
    \end{minipage}
    \captionof{figure}{Column 1: Face images of two celebrities, Dakota Fanning and David Gallagher, obtained from ITWCC Dataset~\cite{ITWCC}, enrolled in the gallery. Columns 2-4: The same celebrities' probe images at different ages (denoted below each photo). As the child grows older, three state-of-the-art face matchers, FaceNet~\cite{facenet}, CosFace~\cite{cosface}, and a commercial-off-the-shelf (COTS) face matcher, fail to match the enrolled image of the same child (highlighted in red). The correct matches are highlighted in green.}
    \label{fig:frontpage}
\end{center}
}]

\begin{abstract}
Given a gallery of face images of missing children, state-of-the-art face recognition systems fall short in identifying a child (probe) recovered at a later age. We propose an age-progression module that can age-progress deep face features output by any commodity face matcher. For time lapses larger than 10 years (the missing child is found after 10 or more years), the proposed age-progression module improves the closed-set identification accuracy of FaceNet from 40\% to 49.56\% and CosFace from 56.88\% to 61.25\% on a child celebrity dataset, namely ITWCC. The proposed method also outperforms state-of-the-art approaches~\cite{look_through_elapse, decorrelated} with a rank-1 identification rate from 94.91\% to 95.91\% on a public aging dataset, FG-NET, and from 99.50\% to 99.58\% on CACD-VS. These results suggest that aging face features enhances the ability to identify young children who are possible victims of child trafficking or abduction. 

\end{abstract}

\section{Introduction}

Human trafficking is one of the most adverse social issues currently faced by countries worldwide. According to the United Nations Children's Fund (UNICEF) and the Inter-Agency Coordination Group against Trafficking (ICAT), 28\% of the identified victims of human trafficking globally are children\footnote{The United Nations Convention on the Rights of the Child defines a child as “a human being below the age of
18 years unless under the law applicable to the child, majority is attained earlier"~\cite{Child}}~\cite{UNICEF}. 
The Wall Street Journal reported in $2012$ that it is estimated that around $8$ million children go missing around the world every year~\cite{WSJ}. Children separated from their parents, such as refugees and migrants, are most vulnerable to trafficking.
According to the FBI, in $2018$ there were $424,066$ NCIC (National Crime Information Center) entries for missing children in the United States~\cite{FBI}. As of 2018, juveniles under the age of 18 account for $34.8$\% of the total active missing records in NCIC~\cite{FBI}. The actual number of missing children is much more than these official statistics as only a limited number of cases are reported because of the fear of traffickers, lack of information, and mistrust of authorities.

Face recognition is perhaps the most promising biometric technology for recovering missing children, since parents and relatives are more likely to have a lost child's face photograph than other biometric modalities such as fingerprint or iris\footnote{Indeed, face is certainly not the only biometric modality for identification
of lost children. Sharbat Gula, first photographed in 1984 (age 12) in a refugee
camp in Pakistan, was later recovered via iris recognition at the age of 30 from a remote part of Afghanistan in 2002~\cite{sherbat}.}.
While Automated Face Recognition (AFR) systems have been able to achieve high identification rates~\cite{cosface, facenet, nist}, their ability to recognize children as they age is still limited.

\begin{figure}
    \centering
    \captionsetup{font=footnotesize}
    \subfloat[]{
    \includegraphics[width=0.475\linewidth]{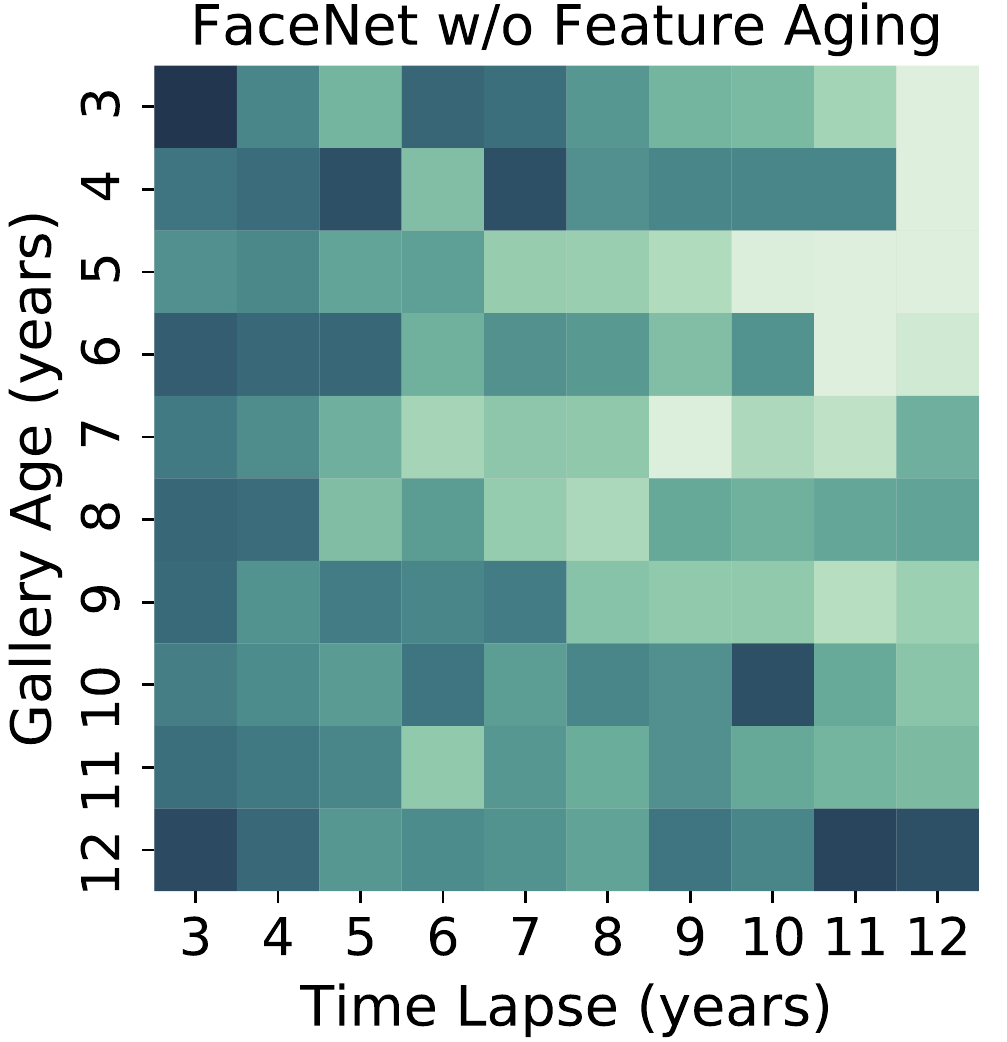}}
    \subfloat[]{
    \includegraphics[width=0.497\linewidth]{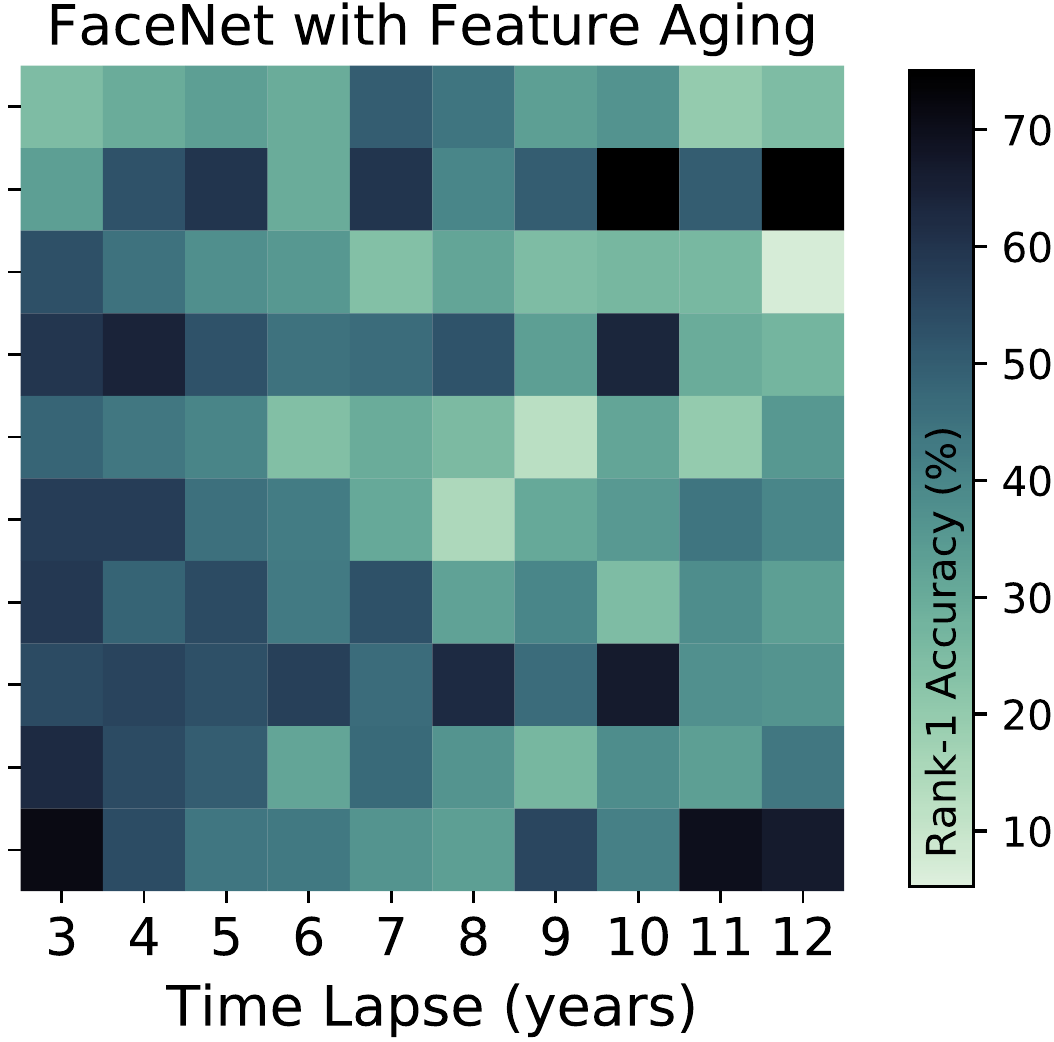}}\vspace{-0.5em}
    \caption{Heat map of rank-1 identification accuracy (\%) (a) without modifying FaceNet features by the proposed aging module and (b) with modifying FaceNet features by the proposed aging module (darker colors indicate higher accuracy). The age of the child in the gallery along with time lapse to the probe are shown along the two axes.}
    \label{fig:heatmap}
\end{figure}

A human face undergoes various temporal changes, including skin texture, weight, facial hair, etc. (see Figure~\ref{fig:frontpage})~\cite{anatomy_face, facialstructure}. Several studies have analyzed the extent to which facial aging affects the performance of AFR (see Table~\ref{tab:related}). Two major conclusions can be drawn based on these studies: (i) Performance decreases with an increase in time lapse between subsequent image acquisitions~\cite{klare,deb_adult,nist_2018}, and (ii) performance degrades more rapidly in the case of younger individuals than older individuals~\cite{nist_2018, deb_child}.
Figure~\ref{fig:heatmap} illustrates that state-of-the-art face matchers fail considerably when it comes to matching an enrolled child in the gallery with the corresponding probe over large time lapses. Thus, it is essential to enhance the longitudinal performance of AFR systems, especially when the child is enrolled at a young age. 

Locating missing children is analogous to the identification scenario (either open-set or closed-set) in face recognition where we search a gallery of missing children to determine the identity of a child retrieved at a later age (probe). As the time gap between a probe image and the true mate in the gallery, gets larger, the search problem gets harder.

\begin{figure}
    \centering
    \captionsetup{font=footnotesize}
    \footnotesize{5 years old\hspace{9.5em} 30 years old}\vspace{-1em}
\subfloat[Saroo Brierley~\cite{saroo_book}\protect\footnotemark]{
    \includegraphics[width=0.505\linewidth]{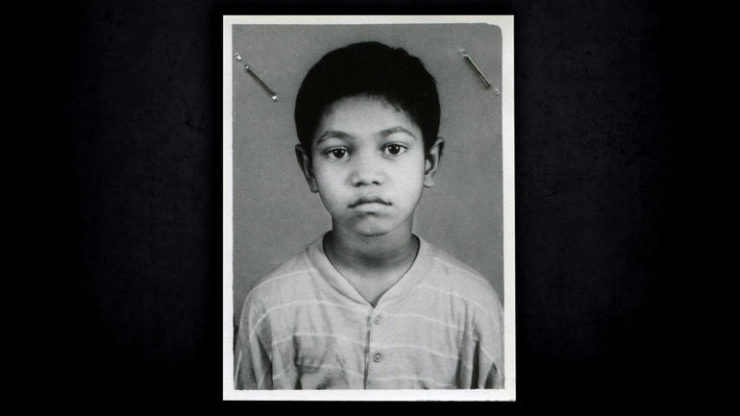}\includegraphics[width=0.4995\linewidth]{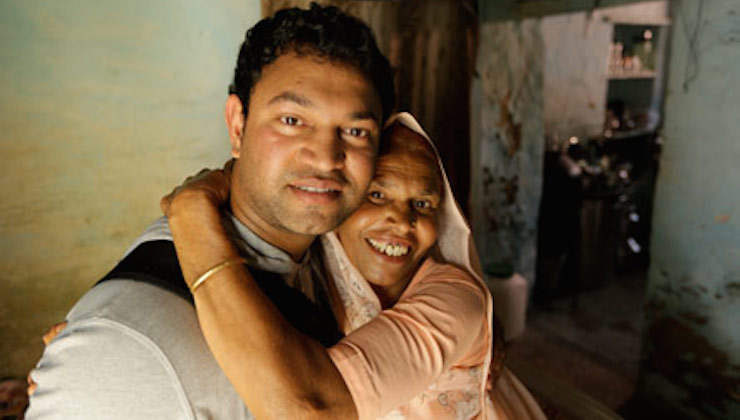}}\vspace{0.5em}
    \footnotesize{11 years old\hspace{9.5em} 29 years old}\vspace{-1em}
    \subfloat[Jaycee Dugard~\cite{stolenlife}]{\includegraphics[width=0.5\linewidth]{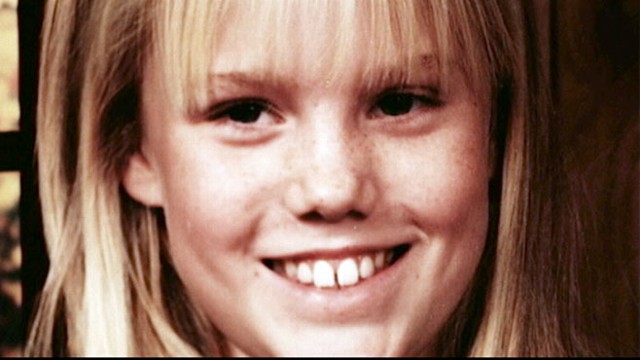}\includegraphics[width=0.5\linewidth]{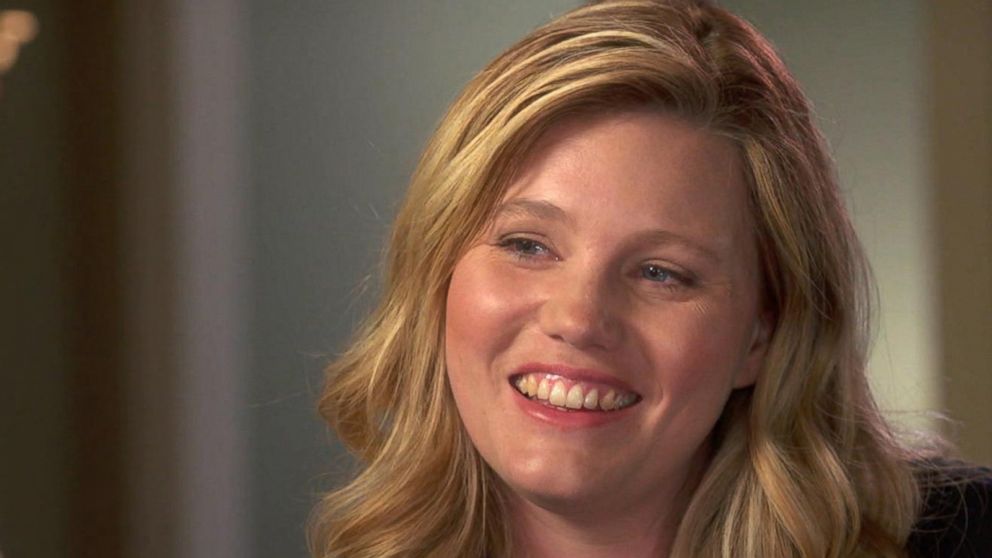}}\vspace{0.5em}
    \footnotesize{19 days old\hspace{9.5em}23 years old}\vspace{-1em}
    \subfloat[\label{fig:carlina_lost}Carlina White~\cite{carlina}]{\hspace{3.5em}\includegraphics[width=0.245\linewidth]{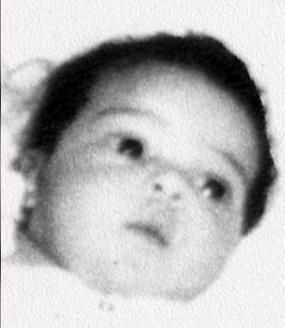}\hspace{3em}\includegraphics[width=0.5\linewidth]{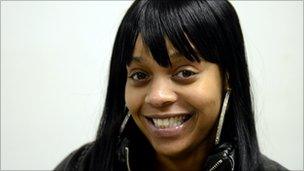}}
    \caption{Face images of missing children in three high profile cases who were successfully recovered after a large time lapse.}
    \label{fig:lost_child}
\end{figure}

\footnotetext{The award-winning 2016 movie, Lion, is based on the true story of Saroo Brierley~\cite{saroo_movie}.}

Prior studies on face recognition under aging, both for adults and children, explored both \emph{generative} and \emph{discriminative} models. Given a probe face image, generative models can generate face images that can either predict how the person will look over time (age progression) or estimate how he looked in previous years (age regression) by utilizing Generative Adversarial Networks (GANs)~\cite{geng, caae, ipcgan, cgan, hongyu, lanitis}. The primary motivation is to enhance the visual quality of the age progressed or regressed face images, rather than enhancing the face recognition performance. On the other hand, discriminative approaches focus on \emph{age-invariant} face recognition under the assumption that age and identity related information can be separated~\cite{lfcnn, discriminative, aecnn, oecnn, look_through_elapse, decorrelated}. By separating age-related components, only the identity-related information is used for face matching. Since age and identity are highly correlated in the feature space, the task of disentangling them from face embeddings is not only difficult but can also be detrimental to AFR performance~\cite{hill,attributes}.

A majority of the prior studies on \emph{cross-age face recognition}\footnote{Face matching or retrieval under aging changes}~\cite{lfcnn,aecnn,oecnn,hongyu,decorrelated, look_through_elapse} evaluate the performance of their models on longitudinal face datasets, such as MORPH ($13,000$ subjects in the age range of $16$-$77$ years) and CACD ($2,000$ subjects in the age range of $16$-$62$ years), which mainly comprise adult face images. Indeed, some benchmark face datasets such as FG-NET ($82$ subjects in the age range of $0$-$69$ years) do include a small number of children, however, the associated protocol is based on matching~\emph{all possible comparisons} for all ages, which does not explicitly provide child-to-adult matching performance.
Moreover, earlier studies employ \emph{cross-sectional} techniques where the temporal performance is analyzed according to differences between age groups~\cite{decorrelated, caae, bereta}.
In cross-sectional or cohort-based approaches, which age groups or time lapses are evaluated is often arbitrary and
varies from one study to another, thereby, making comparisons between studies difficult~\cite{yoon, lacey_adult}. Furthermore, cross-sectional analysis with summary statistics does not investigate whether age-related face recognition performance trends are due to other noise factors such as variations in illumination, expression, and pose. For these reasons, since facial aging is longitudinal by nature, cross-sectional analysis is not the correct model for exploring aggregated effects~\cite{yoon, lacey, nist_irex_report}. The correct model is the longitudinal model that has been utilized for temporal data for fingerprints~\cite{yoon}, face~\cite{lacey, deb_adult} and iris~\cite{nist_irex_report}.

We propose an age-progression module\footnote{Though our module is not strictly restricted to age-progression, we use the word progression largely because in the missing children scenario the gallery would generally be younger than the probe. Our module does both age-progression and age-regression when we benchmark our performance on public datasets.} that learns a projection in the feature space and can be used as a wrapper around any commodity face matcher. Our module can also synthesize the face image corresponding to aged features for a given individual and specified target age. Our empirical results show that the proposed module, based on an encoder-decoder architecture, can enhance the longitudinal face recognition performance of three face matchers (FaceNet~\cite{facenet}, CosFace~\cite{cosface}, and a commercial-off-the-shelf (COTS) matcher) for matching children as they age.

The contributions of the paper can be summarized as follows:
\begin{itemize}[topsep=0.5em, itemsep=-0.1em]
\item A feature aging strategy for traversing the face manifold in the deep feature space, such that the identity of the subject is preserved while only the age component is progressed or regressed in the face embedding.
    \item  Visualizing the aged face from the age-progressed face features, via a decoder, which illustrates that our proposed method can indeed model the age manifold while preserving the identity information.
    \item With the proposed age-progression module, rank-1 identification rates of a state-of-the-art matcher, CosFace~\cite{cosface}, increase from $94.46\%$ to $95.73\%$ on CFA (a child face aging dataset), and $84.69\%$ to $88.45\%$ on ITWCC~\cite{ITWCC-D1} (a child celebrity dataset). In addition, the proposed module boosts accuracies from $94.91\%$ and $99.50\%$ to $95.91\%$ and $99.58\%$ on FG-NET and CACD-VS respectively, which are the two public face aging benchmark datasets~\cite{fgnet, cacd}\footnote{We follow the exact protocols provided with these datasets. We open-source our code for reproducibility: [\textit{url omitted for blind review}].}.
\end{itemize}

\section{Related Work}

\begin{table*}[!t]
\footnotesize
\captionsetup{font=footnotesize}
\caption{Face aging datasets. Datasets below solid line includes longitudinal face images of children.}
\centering
\begin{threeparttable}
\begin{tabularx}{0.885\linewidth}{l l l l l l l l}	
\noalign{\hrule height 1.5pt}
Dataset & No. of Subjects & No. of Images & No. Images / Subject & Age Range (years) & Avg. Age (years)  & Public\protect\footnotemark\\
   \noalign{\hrule height 1pt}
  MORPH-II~\cite{morph} &13,000 & 55,134 & 2-53 (avg. 4.2) &16-77 & 42 & Yes\\
   \hline
    CACD~\cite{cacd} &  2,000 & 163,446   & 22-139 (avg. 81.7) & 16-62 & 31 & Yes\\\noalign{\hrule height 1.2pt}
    FG-NET~\cite{fgnet} & 82 & 1,002 & 6-18 (avg. 12.2) & 0-69 & 16 &  Yes\\
    \hline
    UTKFace~\cite{caae}\tnote{$\dagger$} & N/A & 23,708 & N/A & 0-116 & 33 & Yes\\
    \hline
    ITWCC~\cite{ITWCC} & 745 & 7,990 & 3-37 (avg. 10.7) & 0-32 & 13 & No\tnote{$\dagger\dagger$}\\
    \hline 
    CLF~\cite{deb_child} & 919 & 3,682 & 2-6 (avg. 4.0) & 2-18 & 8 & No\tnote{$\dagger\dagger$}\\ \hline
    CFA & 9,196 & 25,180 & 2-6 (avg. 2.7) & 2-20 & 10 & No\tnote{$\dagger\dagger$}\\
    \hline
\noalign{\hrule height 1.5pt}
\end{tabularx}
\begin{tablenotes}\footnotesize
\item[$\dagger$]\hspace{0.2em} Dataset does not include subject labels; Only a collection of face images along with the corresponding ages.
\item[$\dagger\dagger$] Concerns about privacy issues are making it extremely difficult for researchers to place the child face images in public domain. 
\end{tablenotes}
\end{threeparttable}
\label{tab:allDatasets}
\end{table*}

\begin{table*}[!t]
\scriptsize
\captionsetup{font=footnotesize}
\caption{Related work on cross-age face recognition. Studies below bold line deal with children.}
\centering
\begin{threeparttable}
\renewcommand{\arraystretch}{1.5}
\begin{tabularx}{\textwidth}{>{\centering\bfseries}l>{}X >{\centering}l >{\arraybackslash}X}
\noalign{\hrule height 1.5pt}
Study & Objective & Dataset & Age groups or range (years)\\
   \noalign{\hrule height 1pt}
   Yang \emph{et al.}~\cite{hongyu}\tnote{*} & Age progression of face images & MORPH-II, CACD & 31-40, 41-50, 50+\\
    \hline
   Wang \emph{et al.}~\cite{decorrelated}\tnote{*} & Decomposing age and identity & MORPH-II, FG-NET, CACD & 0-12, 13-18, 19-25, 26-35, 36-45, 46-55, 56-65, 65+\\
    \hline
 Best-Rowden~\etal~\cite{lacey_adult} & Model for change in genuine scores over time & PCSO, MSP & 18-83\\ \noalign{\hrule height 1.2pt}
  Ricanek~\etal~\cite{ITWCC} &  Face comparison of infants to adults & ITWCC & 0-33\\
    \hline
Deb~\etal~\cite{deb_child} & Feasibility study of AFR for children & CLF & 2-18\\
\hline
This study & Aging face features for enhanced AFR for children & CFA, ITWCC, FG-NET, CACD & 0-18\\
\noalign{\hrule height 1.5pt}
\end{tabularx}
\begin{tablenotes}\footnotesize
\item[*] Study uses cross-sectional model (ages are partitioned into age groups) and not the more appropriate longitudinal model~\cite{deb_adult},~\cite{yoon}.
\end{tablenotes}
 \end{threeparttable}
\label{tab:related}
\end{table*}

\subsection{Discriminative Approaches}
Approaches prior to deep learning leveraged robust local descriptors~\cite{hfa, mefa, hierarchical, discriminative, ling} to tackle recognition performance degradation due to face aging. Recent approaches focus on age-invariant face recognition by attempting to discard age-related information from deep face features~\cite{lfcnn, aecnn, coupled, look_through_elapse, decorrelated}. All these methods operate under two critical assumptions: (1) age and identity related features can be disentangled, and (2) the identity-specific feature is adequate for face recognition performance. Several studies, on the other hand, show that age is indeed a major contributor to face recognition performance~\cite{hill,attributes}. Therefore, instead of completely discarding age factors, we exploit the age-related information to progress or regress the deep feature directly to the desired age.

\subsection{Generative Approaches}
Ongoing studies leverage Conditional Auto-encoders and Generative Adversarial Networks (GANs) to synthesize faces by automatically learning aging patterns from face aging datasets~\cite{recurrent, caae, look_through_elapse, hongyu, cgan, ipcgan}. The primary objective of these methods is to synthesize visually realistic face images that appear to be age progressed, and therefore, a majority of these studies do not report the recognition rates.


\subsection{Face Aging for Children}
Best-Rowden~\etal studied face recognition performance of
newborns, infants, and toddlers (ages 0 to 4 years) on 314 subjects acquired over a maximum time lapse of only one
year~\cite{lacey}. Their results showed a True Accept Rate (TAR) of 47.93\% at 0.1\% False Accept Rate (FAR) for an age group of [0, 4] years for a commodity face matcher. Deb~\etal fine-tuned FaceNet~\cite{facenet} to achieve a rank-1 identification accuracy of only $77.86\%$ for a time lapse between the gallery and probe image of 1 year. Srinivas~\etal showed that the rank-1 performance of state of the art commercial face matchers on longitudinal face images from the In-the-Wild Child Celebrity (ITWCC)~\cite{ITWCC-D1} dataset ranges from $44\%$ to $78\%$. These studies (see Table~\ref{tab:related}) primarily focused on evaluating the longitudinal face recognition performance of state-of-the-art face matchers rather than proposing a solution to improve face recognition performance on children as they age. To the best of our knowledge, our study is the first to propose a model for aging deep face features extracted from any commodity face matcher to enhance longitudinal face recognition accuracy on children.

Table~\ref{tab:allDatasets} summarizes longitudinal face datasets that include children and Table~\ref{tab:related} shows related work in this area.
\footnotetext{MORPH: \url{https://bit.ly/31P6QMw}, CACD: \url{https://bit.ly/343CdVd}, FG-NET: \url{https://bit.ly/2MQPL0O}, UTKFace:~\url{https://bit.ly/2JpvX2b}}


\begin{figure}[!t]
\captionsetup{font=footnotesize}
\small
\centering\begin{tabular}{c@{ }c@{ }c@{ }c@{ }c@{ }c@{}}
\textbf{Probe} & \textbf{Age: 10} & \textbf{Age: 30} & \textbf{Age: 70} & \textbf{Age: 90} \\
\includegraphics[width=.19\linewidth]{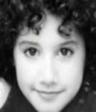}&
\includegraphics[width=.19\linewidth]{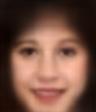}&
\includegraphics[width=.19\linewidth]{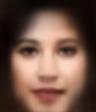}&
\includegraphics[width=.19\linewidth]{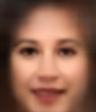}&
\includegraphics[width=.19\linewidth]{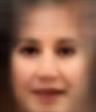}\\
\includegraphics[width=.19\linewidth]{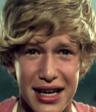}&
\includegraphics[width=.19\linewidth]{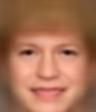}&
\includegraphics[width=.19\linewidth]{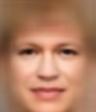}&
\includegraphics[width=.19\linewidth]{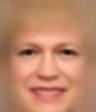}&
\includegraphics[width=.19\linewidth]{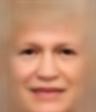}\\
\end{tabular}
\caption{Decoded images from age-progressed features via linear interpolation. The figure indicates that traversing in a linear fashion in the feature space can indeed generate age-progressed features. Here, the two probe images correspond to ages of 5 and 12 years, respectively.}%
\label{fig:mean_features_linear}
\end{figure}

\section{Aging Deep Face Features}\label{sec:tracing_children}

\begin{figure*}
    \centering
    \captionsetup{font=footnotesize}
    \includegraphics[width=\linewidth]{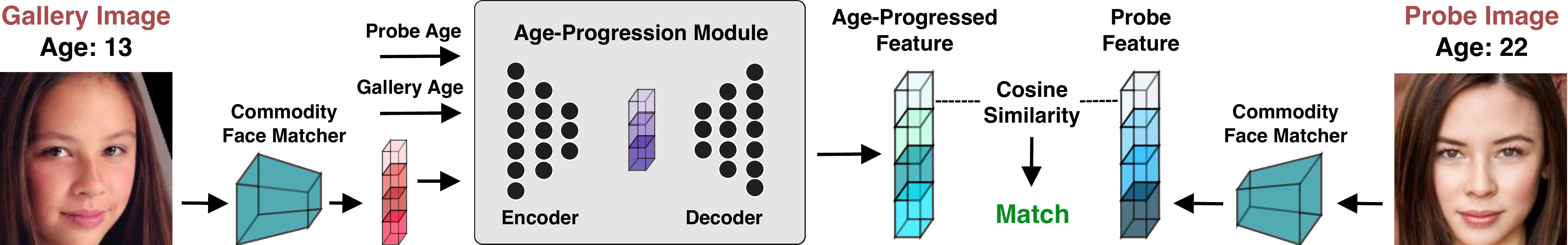}
    \caption{Overview of the proposed deep feature age-progression method. The age-progression module can progress a face feature vector to any desired age.}
    \label{fig:overview}
\end{figure*}

Directly manipulating pixels in a face image for age-progression may not preserve the identity of the child in the \emph{feature space}. Therefore, we propose an age-progression module that learns a projection of the deep features in a lower-dimensional feature space which can directly improve the accuracy of face recognition systems in identifying children over large time lapses (see Figure~\ref{fig:overview}). 

\subsection{Motivation}
 In order to analyze the effect of aging on face matching performance of children, let $\ES = \{\ES^t\}_{t=0}^{T}$, where $T$ is the set of all possible ages in the dataset. Here, $\ES^t = \{\bx_i^t\}_{i=0}^{N_{t}}$, where $\ES^t$ is the set of all $N_t$ images of missing children in the dataset, $\bx_i^t$, acquired at age $t$. Using a commodity face matcher such as FaceNet\footnote{The open-source face matcher, FaceNet, is available at \url{https://github.com/davidsandberg/facenet}.}, we can extract a~\emph{deep feature representation} of the image $\bx_i^t$ which we denote as $\phi(\bx_i^t)$.

We first compute the mean face representation for all ages in $\ES$. For age $t$, we obtain the mean face feature as
\begin{align*}
    \bar{\phi}^{t} = \frac{1}{|\ES^t|}\sum_{i=0}^{N_t} \phi(\bx_i^t) 
\end{align*}
We extract the mean face features $\{\bar\phi^{t}\}_{t=0}^{t=90}$, from the UTKFace dataset~\cite{caae}. In order to isolate the differences induced in the face embeddings due to variations in age, let us define an~\emph{attribute vector} as the difference between any two mean face features at the ages $t_1$ and $t_2$,
\begin{align*}
    \bar\bdelta^{t_1, t_2} = \bar\phi^{t_2} - \bar\phi^{t_1},
\end{align*} where $t_1 << t_2$.
Similar to deep feature interpolation~\cite{deep_feature_interpolation}, we map a child's face image at age $t_1$, $\bx_{i}^{t_1}$, to a point $\phi(\bx_{i}^{t_1})$ in the feature space and move it linearly along the attribute vector $\bar\bdelta^{t_1, t_2}$, via
\begin{align}\label{eq:mean_feat}
    \hat\phi^{t_2} = \phi(\bx_{i}^{t_1}) + \alpha~\bar\bdelta^{t_1, t_2}
\end{align}
In figure~\ref{fig:mean_features_linear}, we show a few decoded image examples when face features for two children, at ages 5 and 12, are moved in this linear fashion along the manifold ($\alpha = 1$).

This experiment indicates that
\begin{itemize}[topsep=0.2em, itemsep=-0.2em]
    \item Face embeddings capture sufficient information about age required for age-progression.
    \item Age-progression can be manifested by \emph{linearly} interpolating in the feature space.
\end{itemize}

\subsection{Learning Feature Age-Progression}
\begin{figure}
    \centering
    \captionsetup{font=footnotesize}
    \includegraphics[width=\linewidth]{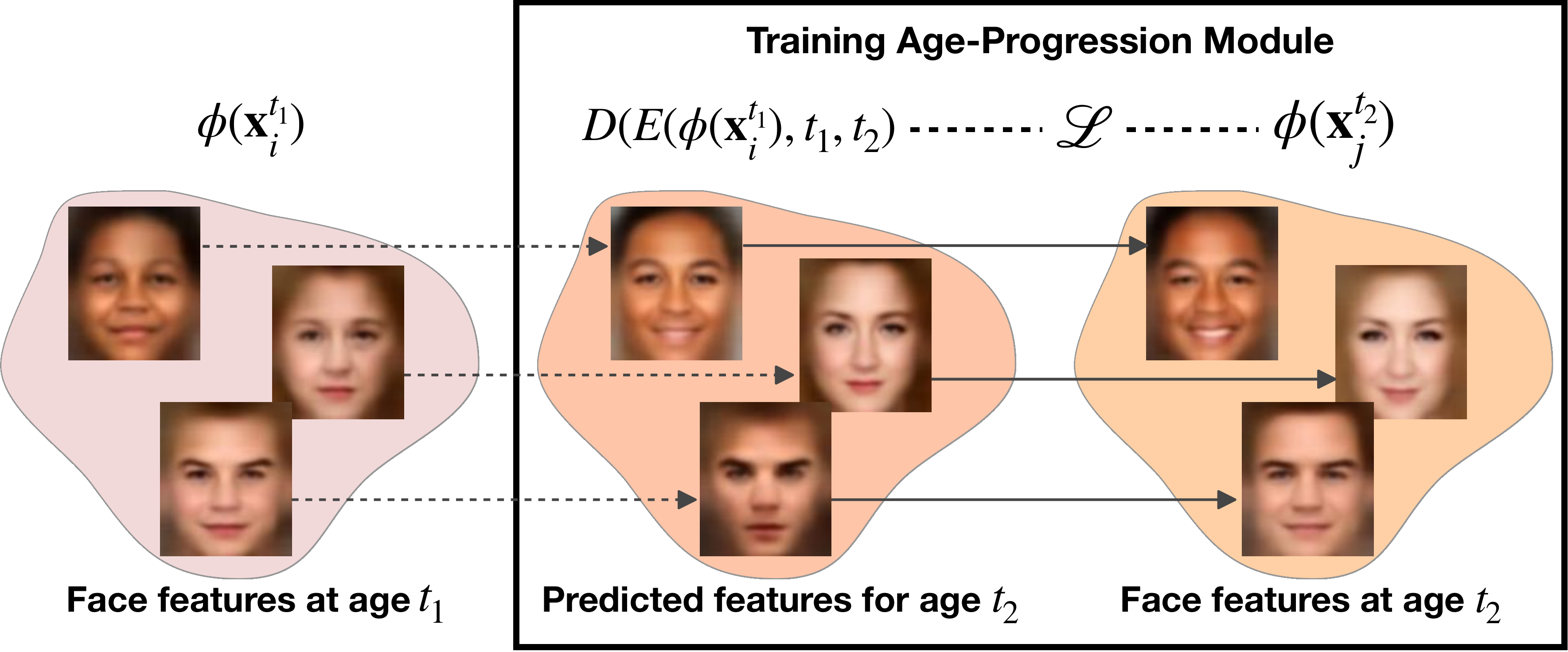}
    \caption{Training the age-progression module. For each face feature acquired at age $t_1$, the predicted features for the desired age, $t_2$, are regressed to the genuine face feature at age $t_2$. }
    \label{fig:training}
\end{figure}

An ideal face feature space $\EZ$ should only encode the identity-salient features and age-related components should be disentangled from identity-relevant features. However, in reality, face matchers naturally encode age-related information in the latent space which has been shown to enhance the discriminative power~\cite{hill,attributes}. We aim to develop an age-progression method that can learn a projection within any face matcher's feature space (see Figure~\ref{fig:training}).

 A pre-trained face matcher embeds a face image, $\bx$ to a $d$-dimensional Euclidean space\footnote{Assume these feature vectors are constrained to lie in a $d$-dimensional hypersphere,~\ie, $||\phi||_2^2 = 1$.}, $\phi(\bx) \in \IR^d$. Assume we have a training set of image pairs, $(\bx_i^{t_1}, \bx_j^{t_2})$, where, $x_i$ and $x_j$ are two images of the same person acquired at ages $t_1$ and $t_2$, respectively. Here, $\bx_i \in \EX, t_a \in \EA$, where $\EX$ is the face image domain and $\EA$ is the set of all possible ages. Our goal is to learn a model that takes a face feature vector, $\phi^{t_1}$, and synthesizes a face embedding for the desired age, $t_2$, such that the identity of the person is preserved while the age-related components are similar to that of $\phi^{t_2}$.

We propose an encoder-decoder architecture that can automatically learn age-progression in the feature space. The encoder $E : (\IR^d, \EA, \EA) \rightarrow \IR^k$ is a stack of fully connected linear layers that maps a feature vector to a $k$-dimensional latent representation $E(\phi(\bx^{t_1}), t_1, t_2)$. The encoder is conditioned on the input feature, $\phi(\bx^{t_1})$, the age at image acquisition, $t_1$, and the desired age after progression, $t_2$. The decoder $D : \IR^k \rightarrow \IR^d$ is also a stack of fully connected linear layers that synthesize an age-progressed version of the original face feature $\phi(\bx^{t_1})$, given its latent representation $E(\phi(\bx^{t_1}), t_1, t_2)$.
In order to ensure that the identity-salient features are preserved and the synthesized features are age-progressed to the desired age, we use train the age-progression module via a mean squared error (MSE) loss which measures the quality of the predicted features:
\begin{align}
    \EL = \frac{1}{|\EP|} \sum_{(i, j)\in \EP} || D(E(\phi(\bx_i^{t_1}), t_1, t_2)) - \phi(\bx_j^{t_2})||_2^2,
\end{align}
where $\EP$ is the set of all genuine pairs. After the model is trained, the age-progression module can progress a face feature to the desired age.

\section{Implementation Details}
\paragraph{Age-Progression Module} The proposed feature aging scheme consists of two components: (a) an encoder and (b) a decoder. For all the experiments, we employ a fully connected linear layer for the encoder and decoder and empirically set the output of each layer to be of the same size as the input feature vector,~\ie, $d = k$. We do not apply any non linear activation function to the fully connected layers. We train the module for 2,000 iterations with a learning rate of $0.1$ using Adam optimizer ($\beta_1 = 0.5, \beta_2 = 0.9$).

\paragraph{Face Matcher} For all our experiments, we employ 3 face matchers\footnote{Both the open-source matchers and the COTS matcher achieve 99\% accuracy on LFW under LFW protocol.}. Two of them, FaceNet~\cite{facenet} and CosFace~\cite{cosface}, are publicly available. FaceNet is trained on VGGFace2 dataset~\cite{vggface2} using the Softmax+Center Loss~\cite{facenet}. CosFace is a 64-layer residual network~\cite{sphereface} and is trained on MS-ArcFace dataset~\cite{arcface} using AM-Softmax loss function~\cite{arcface}. Both matchers extract a $512$-dimensional feature vector. We also evaluate results on a commercial-off-the-shelf (COTS) face matcher, COTS\footnote{This particular COTS utilizes CNNs for face recognition and has been used for identifying children in prior studies~\cite{ITWCC-D1, deb_child}. COTS is one of the top performers in the NIST Ongoing Face Recognition Vendor Test (FRVT)~\cite{nist_2018}.}. This is a closed system so we do not have access to its feature vector.


\begin{table*}[t]
\centering
\footnotesize
\captionsetup{font=footnotesize}
\caption{The mean and standard deviation of face recognition performances across 5-folds on CFA and ITWCC~\cite{ITWCC-D1} datasets with and without proposed deep feature aging. The proposed age-progression method improves the performance of FaceNet and CosFace on cross age face matching.}
\begin{threeparttable}
\begin{tabular}{l||c|c|c||c|c|c}
\noalign{\hrule height 1.5pt}
                & \multicolumn{3}{c||}{\textbf{CFA (Constrained)}}
                & \multicolumn{3}{c}{\textbf{ITWCC (Semi-Constrained)~\cite{ITWCC-D1}}}\\ \hline
\textbf{Method} & \textbf{Verification} & \textbf{Closed-set} & \textbf{Open-set\tnote{$\dagger$}} & \textbf{Verification} & \textbf{Closed-set} & \textbf{Open-set\tnote{$\dagger$}}  \\ 
\noalign{\hrule height 0.1pt}
                & \textbf{0.1\% FAR}               & \textbf{Rank-1 }             & \textbf{Rank-1 @ 0.1\% FAR}            &  \textbf{0.1\% FAR}               &  \textbf{Rank-1}              & \textbf{Rank-1 @ 0.1\% FAR}    \\
\noalign{\hrule height 1.2pt}
COTS~\cite{ITWCC-D1} & 89.96 $\pm$ 4.93 &  95.26 $\pm$ 0.77 & 90.65 $\pm$ 0.59 & 52.23 $\pm$ 7.13 & 86.47 $\pm$ 2.25 & 30.87 $\pm$ 5.35\\\noalign{\hrule height 1.5pt}
FaceNet~\cite{facenet} (w/o feature aging)  & 37.74 $\pm$ 5.16 & 79.42 $\pm$ 0.80 & 49.37 $\pm$ 2.90 & 21.97 $\pm$ 3.60  & 60.53 $\pm$ 2.30 & \textbf{6.30} $\pm$ \textbf{2.30}\\ \hline
FaceNet (with feature aging)  & \textbf{48.92} $\pm$ \textbf{3.93} & \textbf{84.07} $\pm$ \textbf{0.56} & \textbf{56.72} $\pm$ \textbf{0.46} & \textbf{25.72} $\pm$ \textbf{4.41} & \textbf{68.19} $\pm$ \textbf{0.91}  & 5.63 $\pm$ 1.83\\ \noalign{\hrule height 1.5pt}
CosFace~\cite{cosface} (w/o feature aging)   & 81.26 $\pm$ 3.38 & 94.46 $\pm$ 0.25 & 90.23 $\pm$ 0.62 & 49.63 $\pm$ 3.41 & 84.69 $\pm$ 2.09   &  31.27 $\pm$ 5.76 \\ \hline
CosFace (with feature aging)   & \textbf{89.29} $\pm$ \textbf{3.82} & \textbf{95.73} $\pm$ \textbf{0.30} & \textbf{91.89} $\pm$ \textbf{0.98} & \textbf{51.55} $\pm$ \textbf{3.74} & \textbf{88.45} $\pm$ \textbf{1.15} &  \textbf{32.88} $\pm$ \textbf{5.67}\\
\noalign{\hrule height 1.5pt}
\end{tabular}
\begin{tablenotes}\scriptsize
\item[$\dagger$]A probe first claims to be present in the gallery. We accept or reject this claim based on a  pre-determined threshold @ $0.1\%$ FAR (verification). If the probe is accepted, the ranked list of
gallery images which match the probe with similarity scores
above the threshold are returned as the candidate list (identification).
\end{tablenotes}
\end{threeparttable}
\label{tab:results}
\end{table*}

\section{Experimental Results}

\begin{figure}[!t]
\captionsetup{font=footnotesize}
\footnotesize
\centering\begin{tabular}{@{}c@{ }c@{ }c@{ }c@{ }c@{ }}
5 years & 6 years & 8 years & 11 years\\
\includegraphics[width=0.245\linewidth]{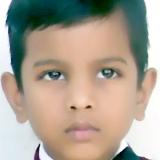}&
\includegraphics[width=0.245\linewidth]{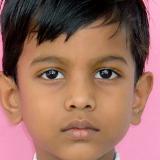}&
\includegraphics[width=0.245\linewidth]{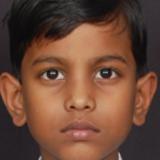}&
\includegraphics[width=0.245\linewidth]{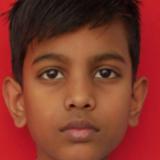}\\
\end{tabular}\\
\textbf{\footnotesize  CFA}\\[0.5em]
\centering\begin{tabular}{@{}c@{ }c@{ }c@{ }c@{ }c@{ }}
3 years & 5 years & 12 years & 13 years\\
\includegraphics[width=0.245\linewidth]{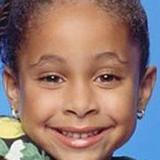}&
\includegraphics[width=0.245\linewidth]{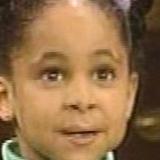}&
\includegraphics[width=0.245\linewidth]{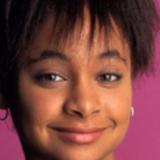}&
\includegraphics[width=0.245\linewidth]{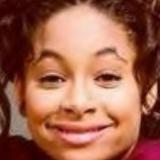}\\\end{tabular}\\
\textbf{\footnotesize ITWCC~\cite{ITWCC-D1}}
\caption{Examples of longitudinal face images from CFA and ITWCC~\cite{ITWCC-D1} datasets. Each row consists of images of one subject; age at image acquisition is given below each image}%
\label{fig:Dataset_examples}
\end{figure}

\begin{figure*}
\centering
\captionsetup{font=footnotesize}
    \subfloat[CFA]{\includegraphics[width=0.4\linewidth]{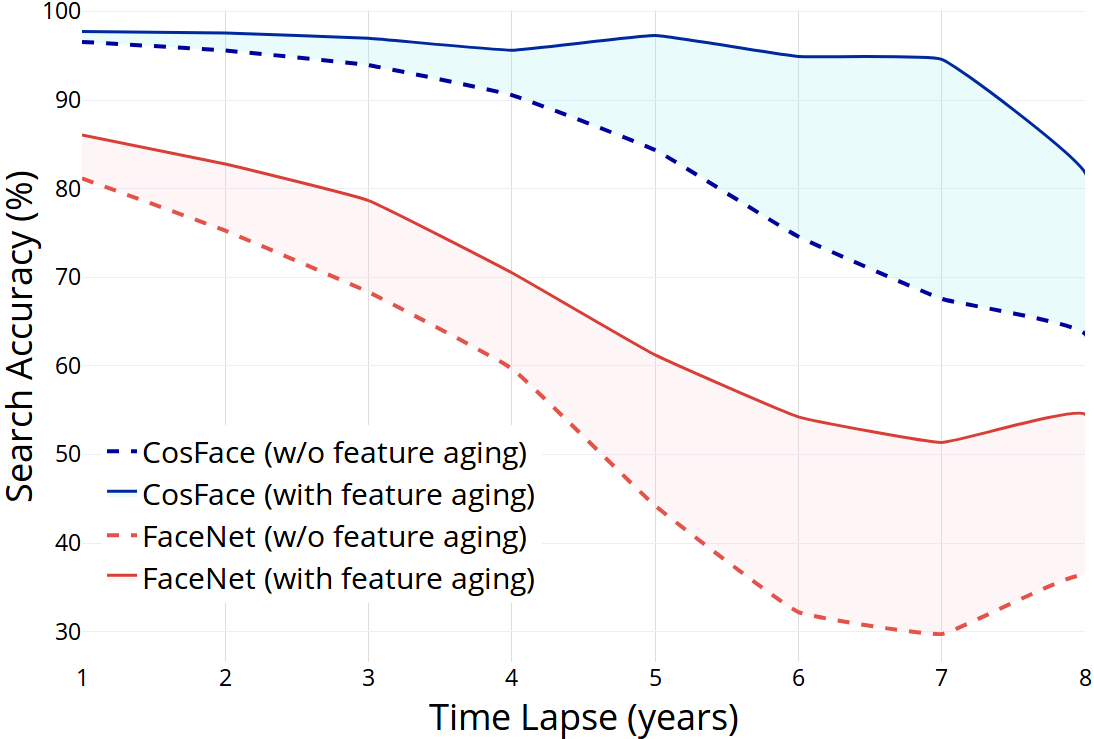}\label{fig:identification_CLF}}\hspace{5em}
    \subfloat[ITWCC]{\includegraphics[width=0.4\linewidth]{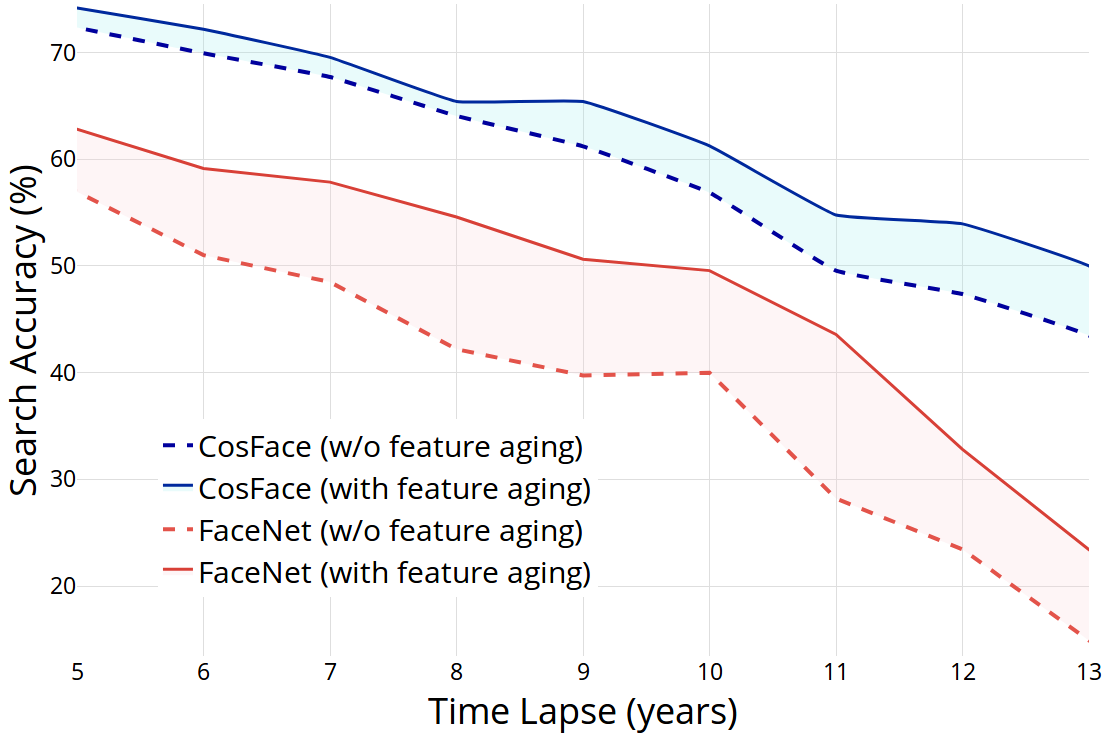}\label{fig:identification_ITWCC}}
    \caption{Rank-1 search accuracy for FaceNet~\cite{facenet} and CosFace~\cite{cosface} on (a) CFA and (b) ITWCC datasets along with our proposed age-progression method.}
    \label{fig:identification_plots}
\end{figure*}

For evaluating child face performance, we utilized two datasets (see Table~\ref{tab:allDatasets}):
\begin{itemize}[topsep=0.2em, itemsep=-0.2em]
    \item \emph{Children's Face Aging (CFA)} dataset comprises $25,180$ annual school portraits of $9,196$ children varying in the age range of $2-20$ years.
    \item \emph{In The Wild Child Celebrity (ITWCC)}~\cite{ITWCC-D1} dataset contains $7,990$ images of $745$ of child celebrities.
\end{itemize}

\paragraph{Evaluation Protocol} For evaluating recognition performance on CFA and ITWCC datasets, we use 5-fold cross validation where the subjects are disjoint in each fold. $4$ folds are used for training and the $5$th for testing. To compensate for the small size of ITWCC dataset, we augment the training data with CFA when evaluating on ITWCC. As locating missing children is akin to the identification scenario, we compute both the \emph{closed-set} identification accuracy (recovered child is in the gallery) at rank-1 and the rank-1 \emph{open-set} identification accuracy (recovered child may or may not be in the gallery) at $0.1\%$ False Accept Rate. Following the protocol outlined in ITWCC~\cite{ITWCC-D1}, we evaluate our results under the \emph{Youngest vs. Oldest} scenario, where the youngest image of a child is enrolled in the gallery and the oldest image of the same child is the probe image. The gallery and probe sets contain $1,739$ and $149$ images each for CFA and ITWCC datasets, respectively. For the open-set identification scenario, we extend the probe set by adding $12,873$ face images of subjects in the age range of $0$ to $32$ years from UTKFace~\cite{caae} dataset. In addition to closed-set and open-set identification results, we also report the verification rate (TAR @ $0.1\%$ FAR).

\paragraph{Results} We report the mean and the standard deviation across all five folds in Table~\ref{tab:results}. We find that our age-progression method improves the search accuracy of both FaceNet~\cite{facenet} and CosFace~\cite{cosface}. In addition, with the proposed feature aging module, an open-source face matcher CosFace~\cite{cosface} can outperform the COTS matcher\footnote{CosFace~\cite{cosface} matcher takes about $1.56$ms to search for a probe in a gallery of $10,000$ images of missing children, while our model takes approximately $27.45$ms (on a GTX 1080 Ti) to search for a probe through the same gallery size.}.

Figures~\ref{fig:identification_CLF} and~\ref{fig:identification_ITWCC} show the performance under the \emph{Youngest vs Oldest} protocol~\cite{ITWCC-D1}. While our aging model improves matching over all time lapses, its contribution gets larger as the time lapse increases. 

In Figure~\ref{fig:retrieval_results}, we show some example cases where CosFace~\cite{cosface}, without the proposed deep feature aging module, retrieves a wrong child from the gallery at rank-1. With the proposed method, we can correctly identify the true mates and retrieve them at rank-1.

\subsection{Comparison with State-of-the-Art}
In order to evaluate the generalizability of our module, we train it on CFA and ITWCC~\cite{ITWCC-D1} datasets and benchmark our performance on a publicly available aging dataset, FG-NET~\cite{fgnet}, which also contains children. We follow the standard leave-one-out protocol~\cite{discriminative, hfa} in Table~\ref{tab:fgnet}. We find that our proposed feature aging module can enhance the performance of CosFace~\cite{cosface}. We also fine-tuned the last layer of CosFace on the same training set, however, the decrease in accuracy (Tab.~\ref{tab:fgnet}) clearly suggests that moving to a new latent space can inhibit the original features. Our module can boost the performance while still operating in the same feature space as the original matcher.

In addition, we also benchmark our performance on an adult aging dataset, CACD-VS\footnote{Since CACD-VS does not have age labels, we use DEX~\cite{dex} (a publicly available age estimator) to estimate the ages.}. However, note that unlike prior studies~\cite{lfcnn,oecnn,look_through_elapse}, we do not fine-tune our model on the CACD-VS dataset. In Table~\ref{tab:cacd}, the proposed feature aging module enhances the performance of CosFace~\cite{cosface} on CACD-VS showing that our model also aids in face recognition under aging for adults.

\begin{table}[t]
\centering
\footnotesize
\captionsetup{font=footnotesize}
\caption{Face recognition performance on FG-NET~\cite{fgnet}.}
\begin{tabular}{l|c}
\noalign{\hrule height 1.5pt}
               \textbf{Method} & \textbf{Rank-1 (\%)}\\ \hline
\noalign{\hrule height 0.1pt}
                HFA~\cite{hfa} &  69.00\%\\
                MEFA~\cite{mefa} &  76.20\%\\
                CAN~\cite{coupled} &  86.50\%\\
                LF-CNN~\cite{lfcnn} &  88.10\%\\
                AIM~\cite{look_through_elapse} &  93.20\%\\
                Wang~\etal~\cite{decorrelated} & 94.50\%\\
                \noalign{\hrule height 0.5pt}
                COTS & 93.61\%\\
                CosFace~\cite{cosface} (w/o feature aging)&  94.91\%\\
                CosFace (finetuned on children) & 93.71\%\\
                \textbf{CosFace (with feature aging)} &  \textbf{95.91\%}\\
\noalign{\hrule height 1.5pt}
\end{tabular}
\label{tab:fgnet}
\end{table}

\begin{table}[t]
\centering
\footnotesize
\captionsetup{font=footnotesize}
\caption{Face recognition performance on CACD-VS~\cite{cacd}.}
\begin{tabular}{l|c}
\noalign{\hrule height 1.5pt}
               \textbf{Method} & \textbf{Accuracy (\%)}\\ \hline
\noalign{\hrule height 0.1pt}
                HFA~\cite{hfa} &  84.40\%\\
                CARC~\cite{cacd} &  87.60\%\\
                LF-CNN~\cite{lfcnn} &  98.50\%\\
                OE-CNN~\cite{oecnn} &  99.20\%\\
                AIM~\cite{look_through_elapse} & 99.38\%\\
                Wang~\etal~\cite{decorrelated} & 99.40\%\\
                \noalign{\hrule height 0.5pt}
                COTS & 99.32\%\\
                CosFace~\cite{cosface} (w/o feature aging)&  99.50\%\\
                \textbf{CosFace (with feature aging)} &  \textbf{99.58\%}\\
\noalign{\hrule height 1.5pt}
\end{tabular}
\label{tab:cacd}
\end{table}

\subsection{Discussion}
\begin{figure}[!t]
\captionsetup{font=footnotesize}
\footnotesize
\centering\begin{tabular}{@{}c@{ }c@{ }c@{ }c@{ }c@{}}
\textbf{Probe: 7 years} & \textbf{Age: 15} & \textbf{Age: 17}  & \textbf{Age: 19} \\
\includegraphics[width=.23\linewidth]{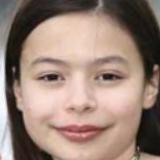}&
\includegraphics[width=.23\linewidth]{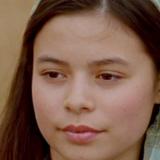}&
\includegraphics[width=.23\linewidth]{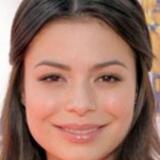}&
\includegraphics[width=.23\linewidth]{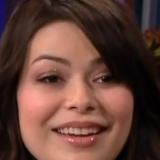}\\
\includegraphics[width=.23\linewidth]{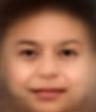}&
\includegraphics[width=.23\linewidth]{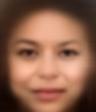}&
\includegraphics[width=.23\linewidth]{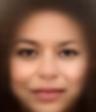}&
\includegraphics[width=.23\linewidth]{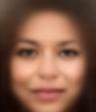}
\end{tabular}
\caption{Face images decoded from age-progressed features (Row 2) along with ground truth face images (ITWCC~\cite{ITWCC-D1}) (Row 1) at target ages.}%
\label{fig:decoded_examples}
\end{figure}

\paragraph{Visualization} Using a decoder, we visualize the aged images from the age-progressed features via the proposed module in Figure~\ref{fig:decoded_examples} for CosFace~\cite{cosface}\footnote{Note that the decoder does not have any explicit age information and it is trained on a non-cross-age face dataset. Implementation details can be found in supplementary material.}. We find that the age-progression module can retain the identity information in the probe's deep face feature while age-related information is progressed as expected.

\begin{figure}
    \centering
    \captionsetup{font=footnotesize}
    \includegraphics[width=0.8\linewidth]{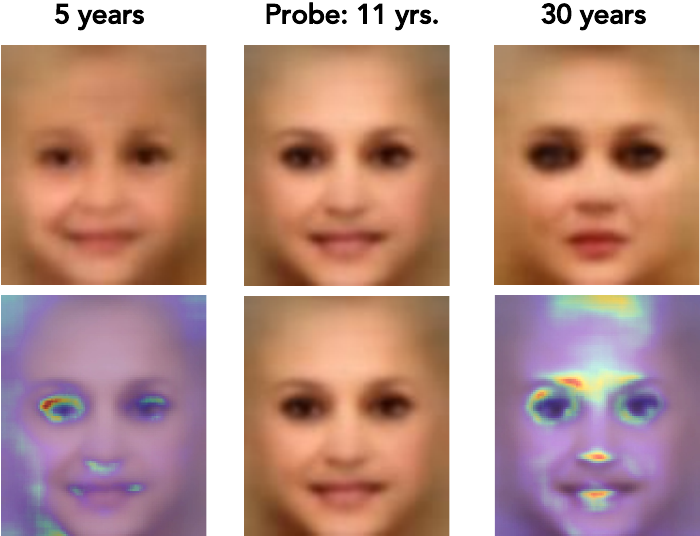}
    \caption{The decoded image for a probe feature at age 11 is shown in column 2. The decoded features at the desired ages via our proposed age-progression module are shown in row 1. Row 2 shows the changes in face features when the probe feature is regressed (column 1) and progressed (column 3). Brighter colors indicate major changes in the face features.}
    \label{fig:perturbation}
\end{figure}

\paragraph{Age-Sensitive Features} 
In Figure~\ref{fig:perturbation}, we visualize the difference between (a) the decoded image from age-progressed or age-regressed features predicted by our proposed method and (b) the decoded image from the probe feature using CosFace~\cite{cosface}. We find that when a probe feature is regressed to a younger age, our method attempts to reduce the size of the head and the eyes, whereas, age-progression enlarges the head, adds makeup, and adds aging effects such as wrinkles around the cheeks.
\section{Two Case Studies of Missing Children}\label{sec:case_study}
Carlina White was abducted from the Harlem hospital center in New York City when she was just $19$ days old. She was reunited with her parents $23$ years later when she saw a photo resembling her as a baby on the National Center for Missing and Exploited Children website\footnote{\url{http://www.missingkids.org}} (see Figure~\ref{fig:carlina_lost}). We contructed a gallery of missing children consisting of $12,873$ face images in the age range $0$ - $32$ years from the UTKFace~\cite{caae} dataset and Carlina's image as an infant when she went missing (19 days old). Her face image when she was later found (23 years old) was used as the probe.
State-of-the-art face matchers, CosFace~\cite{cosface} and COTS, were able to retrieve probe's true mate at ranks $3,069$ and $1,242$ respectively. It is infeasible for a human operator to look through such a large number of retrieved images to ascertain the true mate. With the proposed feature aging module, CosFace is able to retrieve the true mate at \textbf{rank 268}, which is a significant improvement in narrowing down the search.

\begin{figure}[!t]
\footnotesize
\captionsetup{font=footnotesize}
\setlength{\fboxrule}{0.2em}
\centering\begin{tabular}{@{}c@{ }c@{ }c@{ }c@{ }c@{ }}
\textbf{Probe} & \textbf{CosFace}~\cite{cosface} & \textbf{CosFace + Aging}\\[0.2em]
\includegraphics[width=0.25\linewidth]{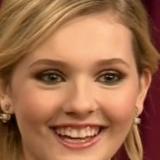}&
\fcolorbox{Red}{white}{\includegraphics[width=0.25\linewidth]{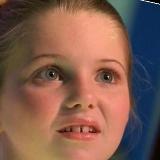}}&
\fcolorbox{Green}{white}{\includegraphics[width=0.25\linewidth]{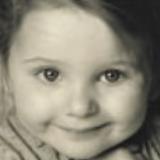}}\\
15 years & 8 years & 2 years\\
\end{tabular}\\\vspace{-1.6em}
\subfloat[\textbf{\footnotesize{ITWCC}}]{~~~~~~~~~~~~~~~~~~~~~~}\vspace{0.5em}
\centering\begin{tabular}{@{}c@{ }c@{ }c@{ }c@{ }c@{ }}
\includegraphics[width=0.25\linewidth]{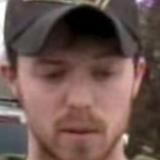}&
\fcolorbox{Red}{white}{\includegraphics[width=0.25\linewidth]{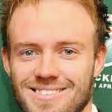}}&
\fcolorbox{Green}{white}{\includegraphics[width=0.25\linewidth]{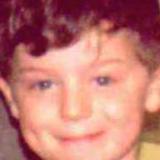}}\\
24 years & 28 years & 5 years
\end{tabular}\\\vspace{-1.6em}
\subfloat[\label{fig:richard_missing}\textbf{\footnotesize Richard Landers: A case study~\cite{michael} (see Section~\ref{sec:case_study})}]{~~~~~~~~~~~~~~~~~~~~~~~~~~~~~~~~~~~~~~~~~~~~~~~~~~~~~~~~~~~~~~~~~~~~~~~~~~~~~~~~~~~~~~~~~~~~~~~~~~~~~~~~~~~~~~}\vspace{0.5em}
\caption{{Identities incorrectly retrieved at Rank-1 by CosFace~\cite{cosface} without our proposed age-progression module (highlighted in red). CosFace with the proposed method can correctly retrieve the true mates in the gallery at rank-1 (highlighted in green)}}%
\label{fig:retrieval_results}
\end{figure}

In another missing child case, Richard Wayne Landers was abducted by his grandparents at age $5$ in July 1994 in Indiana. In 2013, investigators identified Richard (then, $24$ years old) through a Social Security database search (see Figure~\ref{fig:richard_missing}). Akin to Carlina's case, adding Richard's $5$ year old face image in the gallery and keeping his face image at age $24$ as the probe, CosFace~\cite{cosface} was able to retrieve his younger image at rank 23. With the proposed feature aging, CosFace was able to retrieve his younger image at \textbf{rank 1}.

These examples show the applicability of our feature aging module to real world missing children cases. By improving the search accuracy of any face matcher in children-to-adult matching, our model makes a significant contribution to the social good by reuniting missing children with their loved ones.

\section{Summary}
We propose a new method for aging deep face features that can be used as a wrapper around any commodity face matcher to enhance the longitudinal face recognition performance in identifying missing children. The proposed method boosts the rank-1 identification accuracies of FaceNet from 40.00\% to 49.56\% and CosFace from 56.88\% to 61.25\% on a child celebrity dataset, namely, ITWCC. Moreover, with the proposed method, rank-1 accuracy of CosFace on a public aging face dataset, FG-NET, increases from 94.91\% to 95.91\%, outperforming state-of-the-art. These results suggest that aging face features can enhance the ability of commodity face matchers to locate and identify young children who are lost at a young age in order to reuinte them back with their families. We plan to extend our work to unconstrained child face images which is typical in child trafficking cases.

\newpage
{\small
\bibliographystyle{unsrt}
\bibliography{egbib}
}

\end{document}